# A Semantics and Complete Algorithm for Subsumption in the CLASSIC Description Logic

**Alex Borgida** BORGIDA@CS.RUTGERS.EDU
*Department of Computer Science*
*Rutgers University*
*New Brunswick, NJ 08904 U. S. A.*

**Peter F. Patel-Schneider** PFPS@RESEARCH.ATT.COM
*AT&T Bell Laboratories*
*600 Mountain Avenue*
*Murray Hill, NJ 07974 U. S. A.*

## Abstract

This paper analyzes the correctness of the subsumption algorithm used in CLASSIC, a description logic-based knowledge representation system that is being used in practical applications. In order to deal efficiently with individuals in CLASSIC descriptions, the developers have had to use an algorithm that is incomplete with respect to the standard, model-theoretic semantics for description logics. We provide a variant semantics for descriptions with respect to which the current implementation is complete, and which can be independently motivated. The soundness and completeness of the polynomial-time subsumption algorithm is established using description graphs, which are an abstracted version of the implementation structures used in CLASSIC, and are of independent interest.

## 1. Introduction to Description Logics

Data and knowledge bases are models of some part of the natural world. Such models are often built from individual objects that are inter-related by relationships and grouped into classes that capture commonalities among their instances. *Description logics* (DLs), also known as *terminological logics*, form a class of languages used to build and access such models; their distinguishing feature is that classes (usually called *concepts*) can be *defined intensionally*—in terms of descriptions that specify the properties that objects must satisfy to belong to the concept. These descriptions are expressed using some language that allows the construction of composite descriptions, including restrictions on the binary relationships (usually called *roles*) connecting objects.

As an example, consider the description

$$\mathsf{GAME} \sqcap \geq 4\,\mathsf{participants} \sqcap \forall \mathsf{participants}{:}(\mathsf{PERSON} \sqcap \mathsf{gender} : \mathsf{Female}).^{1}$$

This description characterizes objects in the intersection ($\sqcap$) of three sub-descriptions: GAME—objects that belong to the atomic concept; $\geq 4$ participants—objects with at least four fillers for the participants role; and $\forall$participants:(PERSON $\sqcap$ gender : Female)—objects all of whose participants fillers are restricted to belong to PERSONs, which themselves have gender role filled by the value Female.

1. The notation used for descriptions here is the standard notation in the description logic community (Baader et al., 1991). The CLASSIC notation is not used because it is more verbose.

A key difference between DLs and the standard representation formalisms based on First-Order Logic, e.g., relational and deductive databases, is that DLs provide an arena for exploring new sets of "logical connectives"—the constructors used to form composite descriptions—that are different from the standard connectives such as conjunction, universal quantifiers, etc.. Therefore, DLs provide a new space in which to search for expressive yet effectively computable representation languages. Moreover, although it is possible to translate many aspects of DLs currently encountered into First Order Logic, reasoning with such a translation would be a very poor substitute because DL-based systems reason in a way that does not resemble standard theorem proving (e.g., by making use of imperative programming features).

Descriptions such as the one above can be used in several ways in a knowledge base management system (KBMS) based on a description logic:

1. *To state queries:* The KBMS can locate all the objects that satisfy the description's properties.

2. *To define and classify concepts*: Identifiers can be attached to descriptions, in the manner of views in relational DBMSs. The system can in addition automatically determine the "subclass" relationship between pairs of such concepts based on their definitions. For example, a concept defined by the above description would be subsumed by a concept defined by "games with at least two participants" ($\mathsf{GAME} \sqcap \geq 2\,\mathsf{participants}$).

3. *To provide partial information about objects*: It is important to understand that distinct DL descriptions can be ascribed to arbitrary individuals (e.g., "today's game of cards—individual Bgm#467—will have exactly two participants from the following set of three ..., all of whom like tea and rum"). Note that unlike database systems, DL-based KBMSs do not require descriptions to be predefined. This provides considerable power in recording partial knowledge about objects.

4. *To detect errors*: It is possible to determine whether two descriptions are disjoint, whether a description is incoherent or not, and whether ascribing a description to an individual leads to an inconsistency.

Quite a number of KBMSs based on description logics have been built, including CLASSIC (Resnick et al., 1992), LOOM (MacGregor & Bates, 1987), and BACK (Peltason et al., 1987). Such systems have been used in several practical situations, including software information bases (Devanbu et al., 1991), financial management (Mays et al., 1987), configuration management (Owsnicki-Klewe, 1988; Wright et al., 1993), and data exploration. Additional signs that DLs are significant subjects of study are the several recent workshops on DLs (Nebel et al., 1991; Peltason et al., 1991; AAAI, 1992).

### 1.1 On the Tractability and Completeness of DL Implementations

The fundamental operation on descriptions is determining whether one description is more general, or *subsumes*, another, in the sense that any object satisfying the latter would also satisfy the conditions of the former. In parallel with the surge of work on finding tractable yet expressive subsets of first order logic, the DL research community has been investigating the complexity of reasoning with various constructors. The first result in this area (Levesque

& Brachman, 1987) showed that even a seemingly simple addition to a very small language can lead to subsumption determination becoming NP-hard. A more recent, striking pair of results (Patel-Schneider, 1989b; Schmidt-Schauss, 1989) shows that adding the ability to represent equalities of role compositions makes the complexity of the subsumption problem leap from quadratic to *undecidable*.

There are three possible responses to these intractability results:

- Provide an *incomplete implementation* of the DL reasoner, in the sense that there are inferences sanctioned by the standard semantics of the constructors that are not performed by the algorithm. This approach, explicitly adopted by the LOOM system implementers (MacGregor & Bates, 1987), and advocated by some users (Doyle & Patil, 1991), has one major difficulty: how can one describe to users the inferences actually drawn by the implementation so that systems with known properties can be implemented on top of such KBMS? Two solutions to this problem have been suggested: alternative semantic accounts (based on weaker, 4-valued logics, for example) (Patel-Schneider, 1989a), and proof-theoretic semantics (Borgida, 1992).

- Provide a *complete implementation* of a specific DL reasoner, acknowledging that in certain circumstances it may take an inordinate amount of time. This approach, followed in systems such as KRIS (Baader & Hollunder, 1991), has the problem of unpredictability: when will the system "go off into the wild blue yonder"? And of course, in some circumstances this is impossible to even attempt since the reasoning problem is undecidable.

- Carefully devise a *language of limited expressive power* for which reasoning is tractable, and then provide a complete implementation for it. This was the approach chosen by the designers of such languages as KANDOR (Patel-Schneider, 1984) and KRYPTON (Brachman et al., 1983), and is close to the approach in CLASSIC (Borgida et al., 1989).

A hidden difficulty in the second and third approach is to produce an implementation that is correct ("complete") with respect to the semantics. This difficulty is illustrated by the discovery, several years later, that the implementation of KANDOR, as well as CANDIDE (Beck et al., 1989), was in fact incomplete, and its subsumption problem is NP-hard (Nebel, 1988), rather than polynomial, as was claimed; this happened despite the fact that KANDOR is a very "small" language in comparison with other DLs, and its implementation appeared to be evidently correct. To avoid such problems, it is necessary to produce convincing demonstrations that the algorithm is correct; several such proofs have in fact already appeared in the DL literature (e.g., (Patel-Schneider, 1987; Hollunder & Nutt, 1990; Donini et al., 1991)), albeit only for languages that have not seen use in practical applications.

## 1.2 Outline

The CLASSIC 1[2] system is a reasoner based on a moderately complicated DL. It is being used in commercial (Wright et al., 1993) and prototype applications at AT&T, and is made available to academic researchers by AT&T Bell Laboratories.

2. CLASSIC 1 is the first released version of CLASSIC. A new version, CLASSIC 2, with a more expressive DL, has recently been released.

One purpose of this paper is to provide a rigorous formal analysis of the correctness and efficiency for the CLASSIC DL subsumption algorithm.[3] We start by presenting such a result for a subset of the language, which we call Basic CLASSIC. The subsumption algorithm relies on the transformation of descriptions into a data structure, which we call *description graphs*, and which are a generalization of A¨-Kaci's psi-terms (1984). In the process of normalizing such a graph to a canonical form, we remove obvious redundancies and explicate certain implicit facts, encoding in particular the infinite set of inferences that can be drawn from so-called "coreference constraints". The correctness of the subsumption algorithm is demonstrated rigorously by showing how to construct (inductively) a counter-model in case the algorithm returns the answer "no".

Next, we explore the effect of adding individuals to descriptions. We show that, using individuals, one can encode disjunctive information leading to the need to examine combinatorially many possibilities. The CLASSIC implementation is in fact incomplete with respect to the standard semantics. The second contribution of this paper is then a well-motivated, understandable, and small change to the standard semantics that alleviates this problem. We extend the subsumption algorithm and its proof of correctness to deal with individuals under the modified semantics, thereby characterizing in some sense the "incompleteness" of the reasoner.

This paper therefore illustrates all three paradigms described above, albeit in a non-standard manner for the second paradigm, and does so for the first time on a realistic language with significant practical use.

## 2. Basic CLASSIC

Descriptions in Basic CLASSIC are built up from a collection of *atomic* concept names, role names, and attribute names. Roles and attributes are always atomic but descriptions can be built up using operators/constructors such as value restrictions and number restrictions, as we indicate below.

Basic CLASSIC incorporates objects from the host programming language,[4] called *host individuals*, which form a distinct group from *classic individuals*; only the latter can have roles or attributes of their own, the former being restricted to be role or attribute fillers.

The denotational semantics of CLASSIC descriptions starts, as usual, with a domain of values, $\Delta$, subsets of which are extensions for descriptions, while subsets of $\Delta \times \Delta$ are extensions of roles and attributes. This domain is in fact disjointly divided into two *realms*, the *host realm*, $\Delta_H$, containing objects corresponding to host language individuals, and the *classic realm* $\Delta_C$, containing the other objects. Every description, except for THING, which denotes the entire domain has as its extension a subset of either the classic realm or the host realm. (NOTHING denotes the empty set, which is therefore both a classic and host concept.) The extension of a role in a possible world is a relation from the classic realm to the entire domain, while the extension of an attribute is a function from the classic realm into the entire domain.

3. In empirical tests (Heinsohn et al., 1992), CLASSIC has emerged as the fastest of the current DL implementations.
4. A general scheme for incorporating such host objects is described in (Baader & Hanschke, 1991).

Host descriptions are relatively simple: (i) HOST-THING, denoting the entire host realm, $\Delta_H$; (ii) special, pre-defined names corresponding to the types in the host programming language; and (iii) conjunctions of the above descriptions. The descriptions corresponding to the host programming language types have pre-defined extensions and subsumption relationships, mirroring the subtype relationship in the host programming language. This subtype relationship is satisfied in all possible worlds/interpretations. We require that (i) all host concepts have an extension that is either of infinite size or is empty; (ii) that if the extensions of two host concepts overlap, then one must be subsumed by the other, i.e., types are disjoint, unless they are subtypes of each other; and (iii) that a host concept has an infinite number of extra instances than each of its child concepts. (These conditions are needed to avoid being able to infer conclusions from the size of host descriptions.) This allows for host concepts like INTEGER, REAL, COMPLEX, and STRING, but not BOOLEAN or NON-ZERO-INTEGER.

Non-host (classic) descriptions in Basic CLASSIC are formed according to the following syntax:

| Syntax | Constructor Name |
|---|---|
| CLASSIC-THING | |
| E | Atomic Concept Name |
| $C \sqcap D$ | Intersection |
| $\forall R{:}C$ | Role Value Restriction |
| $\forall A{:}C$ | Attribute Value Restriction |
| $\geq n\ R$ | Minimum Number Restriction |
| $\leq m\ R$ | Maximum Number Restriction |
| $A_1 \circ \ldots \circ A_k = B_1 \circ \ldots \circ B_h$ | Equality Restriction |

where E is an atomic concept name; C and D are classic descriptions; R is a role; A, $A_i$, and $B_j$ are attributes; n,k,h are positive integers; and m is a non-negative integer. The set of constructors in Basic CLASSIC was judiciously chosen to result in a language in which subsumption is easy to compute.

The denotational semantics for descriptions in Basic CLASSIC is recursively built on the extensions assigned to atomic names by a possible world:

**Definition 1** *A possible world/interpretation, $\mathcal{I}$, consists of a domain, $\Delta$, and an interpretation function $\cdot^{\mathcal{I}}$. The domain is disjointly divided into a classic realm, $\Delta_C$, and a host realm, $\Delta_H$. The interpretation function assigns extensions to atomic identifiers as follows:*

- *The extension of an atomic concept name E is some subset $E^{\mathcal{I}}$ of the classic realm.*
- *The extension of an atomic role name R is some subset $R^{\mathcal{I}}$ of $\Delta_C \times \Delta$.*
- *The extension of an atomic attribute name A is some total function $A^{\mathcal{I}}$ from $\Delta_C$ to $\Delta$.*

*The extension $C^{\mathcal{I}}$ of a non-atomic classic description is computed as follows:*

- $\text{CLASSIC-THING}^{\mathcal{I}} = \Delta_C$.
- $(C \sqcap D)^{\mathcal{I}} = C^{\mathcal{I}} \cap D^{\mathcal{I}}$.

- $(\forall \mathsf{p}{:}\mathsf{C})^{\mathcal{I}} = \{d \in \Delta_C \mid \forall x\ (d,x) \in \mathsf{p}^{\mathcal{I}} \Rightarrow x \in \mathsf{C}^{\mathcal{I}}\}$, *i.e., those objects in* $\Delta_C$ *all of whose* p*-role or* p*-attribute fillers are in the extension of* C*;*

- $(\geq \mathsf{n}\ \mathsf{p})^{\mathcal{I}}$ *(resp.* $(\leq \mathsf{n}\ \mathsf{p})^{\mathcal{I}}$*) is those objects in* $\Delta_C$ *with at least (resp. at most)* n *fillers for role* p.

- $(\mathsf{A}_1 \circ \ldots \circ \mathsf{A}_k = \mathsf{B}_1 \circ \ldots \circ \mathsf{B}_h)^{\mathcal{I}} = \{d \in \Delta_C \mid \mathsf{A}_k{}^{\mathcal{I}}(\ldots \mathsf{A}_1{}^{\mathcal{I}}(d)) = \mathsf{B}_h{}^{\mathcal{I}}(\ldots \mathsf{B}_1{}^{\mathcal{I}}(d))\}$, *i.e., those objects in* $\Delta_C$ *with the property that applying the composition of the extension of the* $\mathsf{A}_i$*s and the composition of the extension of the* $\mathsf{B}_j$*s to the object both result in the same value.*[5]

*A description,* $\mathsf{D}_1$*, is then said to* subsume *another,* $\mathsf{D}_2$*, if for all possible worlds* $\mathcal{I}$*,* $\mathsf{D}_2{}^{\mathcal{I}} \subseteq \mathsf{D}_1{}^{\mathcal{I}}$.

Of key interest is the computation of the subsumption relationship between descriptions in Basic CLASSIC. Subsumption computation is a multi-part process. First, descriptions are turned into description graphs. Next, description graphs are put into canonical form, where certain inferences are explicated and other redundancies are reduced by combining nodes and edges in the graph. Finally, subsumption is determined between a description and a canonical description graph.

To describe in detail the above process, we start with a formal definition of the notion of description graph (Definition 2), and then present techniques for

- translating a description to a description graph (Section 2.2), which requires merging pairs of nodes, and pairs of graphs (Definitions 4 and 5);

- putting a description graph into canonical form (Section 2.3);

- determining whether a description subsumes a description graph (Algorithm 1).

To prove the correctness of this approach, we need to show that the first two steps lead us in the right direction, i.e., that the following three questions are equivalent: "Does description D subsume description C?", "Does description D subsume graph $G_C$?", and "Does description D subsume graph *canonical*($G_C$)?". To do this, we need to define the formal semantics of both descriptions and graphs (Definitions 1 and 3), and then prove the results (Theorems 1 and 2). To prove the "completeness" of the subsumption algorithm, we show that if the algorithm does not indicate that D subsumes *canonical*($G_C$), then we can construct an interpretation ("graphical world") in which some object is in the denotation of *canonical*($G_C$) but not that of D.

## 2.1 Description Graphs

One way of developing a subsumption algorithm is to first transform descriptions into a canonical form, and then determine subsumption relationships between them. Canonical descriptions can normally be thought of as trees since descriptions are terms in a first order term language. The presence of equality restrictions in CLASSIC significantly changes the

5. Note that both attribute chains must have a definite value, and that all but the last cannot evaluate to host individuals, since these cannot have attributes.

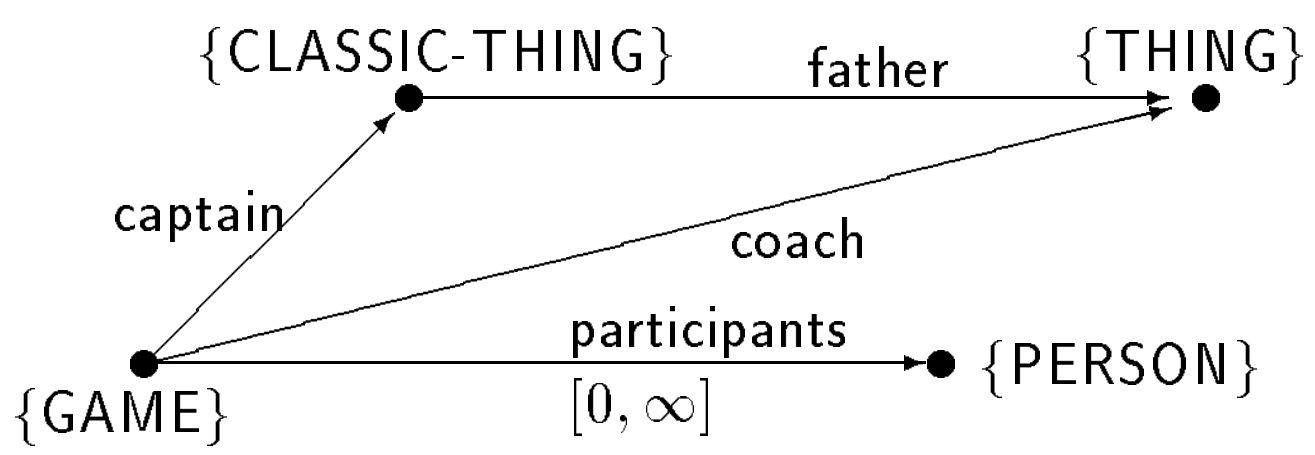

Figure 1: A description graph.

handling of subsumption because they introduce relationships between different pieces of the normal form. Most significantly, in the presence of equalities, a small description, such as $\forall$friend:TALL $\sqcap$ friend = friend$\circ$friend, can be subsumed by descriptions of arbitrary size, such as

$$\forall\text{friend:}(\forall\text{friend:}(\ldots(\forall\text{friend:TALL})\ldots)).$$

In order to record such sets of inferences in the canonical form, we will resort to a graph-based representation, suggested by the semantic-network origins of description logics, and the work of A¨-Kaci (1984).

Intuitively, a *description graph* is a labelled, directed multigraph, with a distinguished node. Nodes of the graph correspond to descriptions, while edges of the graph correspond to restrictions on roles or attributes. The edges of the graph are labelled with the role name and the minimum and maximum number of fillers associated with the edge, or just with the attribute name. The nodes of the graph are labelled with concept names associated with the node concept. For example, Figure 1 is a description graph, which, as we shall see later, corresponds to the description GAME $\sqcap$ $\forall$participants:PERSON $\sqcap$ coach = (captain$\circ$father).

Because equality restrictions (and hence the non-tree portions of the graph) involve only attributes, edges labelled with roles are all cut-edges, i.e., their removal increases by one the number of connected components of the graph. This restriction is important because if the graph is in tree form, there is really no difference between a graphical and a linear notation, and a semantics is simple to develop. If the graph is a general directed acyclic graph, then there is the problem of relating the semantics generated by two different paths in the graph that share the same beginning and ending nodes. If the graph contains cycles, the problem of developing a correct semantics is even more difficult, as a simplistic semantics will be non-well-founded, and some sort of fixed-point or model-preference semantics will be required. Fortunately, any non-tree parts of our graphical notation will involve attributes only, and because attributes are functional, our job will be much easier.

As a result of the above restrictions, it is possible to view a description graph as having the following recursive structure: (i) There is a distinguished node $r$, which has an "island" of nodes connected to it by edges labelled with attributes. (ii) Nodes in this island may have 0 or more edges labelled with roles leaving them, pointing to distinguished nodes of other description graphs. (iii) These graphs share no nodes or edges in common with each other, nor with the islands above them.

Because of this recursive structure, it is easier to represent description graphs using a recursive definition, instead of the usual graph definition. This recursive definition is similar to the recursive definition of a tree, which states that a tree consists of some information (the information on the root of the tree) plus a set of trees (the children of the root of the tree). As description graphs are more complex than simple trees, we will have to use a two-part definition.

**Definition 2** *A* description graph *is a triple, $\langle N, E, r\rangle$, consisting of a set $N$ of* nodes*; a bag $E$ of edges (*a-edges*) labelled with attribute names; and a* distinguished node *$r$ in $N$. Elements of $E$ will be written $\langle n_1, n_2, \mathsf{A}\rangle$ where $n_1$ and $n_2$ are nodes and $\mathsf{A}$ is an attribute name.*

*A* node *in a description graph is a pair, $\langle C, H\rangle$ consisting of a set $C$ of concept names (the* atoms *of the node), and a bag $H$ of tuples (the* r-edges *of the node). An r-edge is a tuple, $\langle \mathsf{R}, m, M, G\rangle$, of a role name, $\mathsf{R}$; a min, $m$, which is a non-negative integer; a max, $M$, which is a non-negative integer or $\infty$; and a (recursively nested) description graph $G$, representing the restriction on the fillers of the role. ($G$ will often be called the* restriction graph *of the node.)*

Concept names *in a description graph are atomic concept names, host concept names,* THING*,* CLASSIC-THING*, or* HOST-THING*.*

Descriptions graphs are provided extensions starting from the same possible worlds $\mathcal{I}$ as used for descriptions. However, in addition we need a way of identifying the individuals to be related by attributes, which will be given by the function $\Upsilon$.

**Definition 3** *Let $G = \langle N, E, r\rangle$ be a description graph and let $\mathcal{I}$ be a possible world. Then the interpretation $G^{\mathcal{I}}$ of $G$, and the interpretation $n^{\mathcal{I}}$ of each of the nodes in $N$, are recursively (and mutually) defined as follows:*

*An element, $d$, of $\Delta$ is in $G^{\mathcal{I}}$, iff there is some function, $\Upsilon$, from $N$ into $\Delta$ such that*

1. *$d = \Upsilon(r)$;*
2. *for all $n \in N$ $\Upsilon(n) \in n^{\mathcal{I}}$;*
3. *for all $\langle n_1, n_2, \mathsf{A}\rangle \in E$ we have $\langle \Upsilon(n_1), \Upsilon(n_2)\rangle \in \mathsf{A}^{\mathcal{I}}$, (which is equivalent to $\Upsilon(n_2) = \mathsf{A}^{\mathcal{I}}(\Upsilon(n_1))$, since $\mathsf{A}^{\mathcal{I}}$ is a function).*

*An element, $d$, of $\Delta$ is in $n^{\mathcal{I}}$, where $n = \langle C, H\rangle$, iff*

1. *for all $\mathsf{C} \in C$, we have $d \in \mathsf{C}^{\mathcal{I}}$; and*
2. *for all $\langle \mathsf{R}, m, M, G\rangle \in H$,*
   (a) *there are between $m$ and $M$ elements, $d'$, of the domain such that $\langle d, d'\rangle \in \mathsf{R}^{\mathcal{I}}$ and*
   (b) *$d' \in G^{\mathcal{I}}$ for all $d'$ such that $\langle d, d'\rangle \in \mathsf{R}^{\mathcal{I}}$.*

## 2.2 Translating Descriptions to Description Graphs

A Basic CLASSIC description is turned into a description graph by a recursive process, working from the "inside out". In this process, description graphs and nodes are often merged.

**Definition 4** *The merge of two nodes, $n_1 \oplus n_2$, is a new node whose atoms are the union of the atoms of the two nodes and whose r-edges are the union of the r-edges of the two nodes*[6]*.*

**Definition 5** *The merge of two description graphs, $G_1 \oplus G_2$, is a description graph whose nodes are the disjoint union*[7] *of the non-distinguished nodes of $G_1$ and $G_2$ plus a new distinguished node. The edges of the merged graph are the union of the edges of $G_1$ and $G_2$, except that edges touching on the distinguished nodes of $G_1$ or $G_2$ are modified to touch the new distinguished node. The new distinguished node is the merge of the two distinguished nodes of $G_1$ and $G_2$.*

The rules for translating a description C in Basic CLASSIC into a description graph $G_{\mathsf{C}}$ are as follows:

1. A description that consists of a concept name is turned into a description graph with one node and no a-edges. The atoms of the node contains only the concept name. The node has no r-edges.

2. A description of the form $\geq$n R is turned into a description graph with one node and no a-edges. The node has as its atoms CLASSIC-THING and has a single r-edge with role R, min $n$, max $\infty$, and restriction $G_{\mathsf{THING}}$.

3. A description of the form $\leq$n R is turned into a description graph with one node and no a-edges. The node has as its atoms CLASSIC-THING and a single r-edge with role R, min 0, max $n$, and restriction $G_{\mathsf{THING}}$.

4. A description of the form $\forall$R:C, with R a role, is turned into a description graph with one node and no a-edges. The node has as its atoms CLASSIC-THING and has a single r-edge with role R, min 0, max $\infty$, and restriction $G_{\mathsf{C}}$.

5. To turn a description of the form C $\sqcap$ D into a description graph, construct $G_{\mathsf{C}}$ and $G_{\mathsf{D}}$ and merge them.

6. To turn a description of the form $\forall$A:C, with A an attribute, into a description graph, first construct the description graph $\langle N_C, E_C, r_C \rangle$ for C. The description graph for $\forall$A:C is $\langle N_C \cup \{t\}, E_C \cup \{\langle t, r_C, \mathsf{A}\rangle\}, t\rangle$, where $t$ is the node $\langle\{\text{CLASSIC-THING}\}, \{\}\rangle$.

7. To turn a description of the form $\mathsf{A}_1 \circ \ldots \circ \mathsf{A}_n = \mathsf{B}_1 \circ \ldots \circ \mathsf{B}_m$ into a description graph first create a distinguished node, node $r$, with CLASSIC-THING as its atoms, and a node $e$, with THING as its atoms. For $1 \leq i \leq n-1$ create a node $a_i$, with its atoms

6. Note that duplicate edges, such as ones joining $n_i$ to $n_i$, are not removed, since the edges form a bag.
7. In taking the disjoint union of two sets, elements of one may be systematically renamed first to make sure that the sets are non-overlapping.

being CLASSIC-THING. For $1 \leq j \leq m-1$ create a node $b_j$, with its atoms being CLASSIC-THING. None of the $a_i$ or $b_j$ have r-edges.

If $n = 1$, create the edge $\langle r, e, \mathsf{A_1} \rangle$; if $n > 1$ then create edges $\langle r, a_1, \mathsf{A_1} \rangle$, $\langle a_{n-1}, e, \mathsf{A_n} \rangle$, and $\langle a_{i-1}, a_i, \mathsf{A_i} \rangle$ for $2 \leq i \leq n-1$.

Similarly, if $m = 1$, create the edge $\langle r, e, \mathsf{B_1} \rangle$; if $m > 1$ then create edges $\langle r, b_1, \mathsf{B_1} \rangle$, $\langle b_{m-1}, e, \mathsf{B_m} \rangle$, and $\langle b_{i-1}, b_i, \mathsf{B_i} \rangle$ for $2 \leq i \leq m-1$.

This creates two disjoint paths, one for the $\mathsf{A_i}$ and one for the $\mathsf{B_j}$, from the distinguished node to the end node.

Figure 1 presents a view of a description graph constructed in this fashion from the description GAME $\sqcap$ $\forall$participants:PERSON $\sqcap$ coach = captain$\circ$father.

Now we want to show that this process preserves extensions. As we use the merge operations we first show that they work correctly.

**Lemma 1** *If $n_1$ and $n_2$ are nodes then $(n_1 \oplus n_2)^{\mathcal{I}} = n_1^{\mathcal{I}} \cap n_2^{\mathcal{I}}$. If $D_1$ and $D_2$ are description graphs then $(D_1 \oplus D_2)^{\mathcal{I}} = D_1^{\mathcal{I}} \cap D_2^{\mathcal{I}}$.*

**Proof:** Since the components (atoms and r-edges) of the merged node are obtained by unioning the components of the respective nodes, and since the interpretation of a node is the intersection of the interpretation of its components, the result is obviously true for nodes.

For merging graphs, the only difference is that the root nodes are replaced by their merger in all edges, as well as the root. But then an element of $(D_1 \oplus D_2)^{\mathcal{I}}$ is clearly an element of both $D_1^{\mathcal{I}}$ and $D_2^{\mathcal{I}}$. Conversely, since we take the *disjoint* union of the other nodes in the two graphs, the mapping functions $\Upsilon_1$ and $\Upsilon_2$ in Definition 3 can simply be unioned, so that an element of both $D_1^{\mathcal{I}}$ and $D_2^{\mathcal{I}}$ is an element of the merged root node, and hence of $(D_1 \oplus D_2)^{\mathcal{I}}$. ∎

**Theorem 1** *For all possible worlds, the extension of a description is the same as the extension of its description graph.*

**Proof:** The proof is by structural induction on descriptions.

The extension of concept names, cardinality restrictions, and $\forall$-restrictions on roles can be easily seen to agree with the extension of description graphs formed from them. Lemma 1 shows that conjunction is properly handled. For $\forall$-restrictions on attributes, the construction is correct because attributes are functional.

For equalities $\mathsf{A_1} \circ \ldots \circ \mathsf{A_n} = \mathsf{B_1} \circ \ldots \circ \mathsf{B_m}$ the construction forms a description graph with two disjoint paths from the distinguished node to an end node, one labelled by the $\mathsf{A_i}$, through nodes $a_i$, and the other labelled by the $\mathsf{B_j}$, through nodes $b_j$. If

$$d \in (\mathsf{A_1} \circ \ldots \circ \mathsf{A_n} = \mathsf{B_1} \circ \ldots \circ \mathsf{B_m})^{\mathcal{I}} = \{d \in \Delta_C \mid \mathsf{A_k}^{\mathcal{I}}(\ldots \mathsf{A_1}^{\mathcal{I}}(d)) \ = \ \mathsf{B_h}^{\mathcal{I}}(\ldots \mathsf{B_1}^{\mathcal{I}}(d))\},$$

then defining $\Upsilon(a_i) = \mathsf{A_i}^{\mathcal{I}}(\ldots \mathsf{A_1}^{\mathcal{I}}(d))$ and $\Upsilon(b_j) = \mathsf{B_j}^{\mathcal{I}}(\ldots \mathsf{B_1}^{\mathcal{I}}(d))\}$, yields the mapping required by Definition 3. The converse is satisfied by the requirement in Definition 3 that for each a-edge $\langle n_1, n_2, \mathsf{A} \rangle \in E$, we have $\Upsilon(n_2) = \mathsf{A}^{\mathcal{I}}(\Upsilon(n_1))$. ∎

### 2.3 Canonical Description Graphs

In the following sections we will occasionally refer to *"marking a node incoherent"*; this consists of replacing it with a special node having no outgoing r-edges, and including in its atoms NOTHING, which always has the empty interpretation. Marking a description graph as incoherent consists of replacing it with a description graph consisting only of an incoherent node. (Incoherent graphs are to be thought of as representing concepts with empty extension.)

Description graphs are transformed into canonical form by repeating the following normalization steps whenever possible for the description graph and all its descendants.

1. If some node has in its atoms a pre-defined host concept, add HOST-THING to its atoms. If some node has an atomic concept name in its atoms, add CLASSIC-THING to its atoms. For each pre-defined host concept in the atoms of the node, add all the more-general pre-defined host concepts to its atoms.

2. If some node has both HOST-THING and CLASSIC-THING in its atoms, mark the node incoherent. If some node has in its atoms a pair of host concepts that are not related by the pre-defined subsumption relationship, mark the node incoherent, since their intersection will be empty.

3. If any node in a description graph is marked incoherent, mark the description graph as incoherent. *(Reason:* Even if the node is not a root, attributes must always have a value, and this value cannot belong to the empty set.*)*

4. If some r-edge in a node has its min greater than its max, mark the node incoherent.

5. If some r-edge in a node has its description graph marked incoherent, change its max to 0. *(Reason:* It cannot have any fillers that belong to the empty set.*)*

6. If some r-edge in a node has a max of 0, mark its description graph as incoherent. *(Reason:* This normalization step records the equivalence between $\leq 0$ R and $\forall$R:NOTHING, and is used then to infer that a concept with $\forall$R:C for arbitrary C subsumes $\leq 0$ R.*)*

7. If some node has two r-edges labelled with the same role, merge the two edges, as described below.

8. If some description graph has two a-edges from the same node labelled with the same attribute, merge the two edges.

To *merge two r-edges* of a node, which have identical roles, replace them with one r-edge. The new r-edge has the role as its role, the maximum of the two mins as its min, the minimum of the two maxs as its max, and the merge of the two description graphs as its restriction.

To *merge two a-edges* $\langle n, n_1, \mathsf{A}\rangle$ and $\langle n, n_2, \mathsf{A}\rangle$, replace them with a single new edge $\langle n, n', \mathsf{A}\rangle$, where $n'$ results from merging $n_1$ and $n_2$, i.e., $n' = n_1 \oplus n_2$. (If $n_1 = n_2$ then $n' = n_1$.) In addition, replace $n_1$ and $n_2$ by $n'$ in all other a-edges of this description graph.

We need to show that the transformations to canonical form do not change the extension of the graph. The main difficulty is in showing that the two edge-merging processes do not change the extension.

**Lemma 2** *Let $G = \langle N, E, r\rangle$ be a description graph with two mergeable a-edges and let $G' = \langle N', E', r'\rangle$ be the result of merging these two a-edges. Then $G^{\mathcal{I}} = G'^{\mathcal{I}}$.*

**Proof:** Let the two edges be $\langle n, n_1, \mathsf{A}\rangle$ and $\langle n, n_2, \mathsf{A}\rangle$ and the new node $n'$ be $n_1 \oplus n_2$.

Choose $d \in G^{\mathcal{I}}$, and let $\Upsilon$ be a function from $N$ into the domain satisfying the conditions for extensions (Definition 3) such that $\Upsilon(r) = d$. Then $\Upsilon(n_1) = \Upsilon(n_2)$ because both are equal to $\mathsf{A}^{\mathcal{I}}(\Upsilon(n))$. Let $\Upsilon'$ be the same as $\Upsilon$ except that $\Upsilon'(n') = \Upsilon(n_1) = \Upsilon(n_2)$. Then $\Upsilon'$ satisfies Definition 3, part 3, for $G'$, because we replace $n_1$ and $n_2$ by $n'$ everywhere. Moreover, $\Upsilon'(n') = \Upsilon(n_1) \in n_1^{\mathcal{I}} \cap n_2^{\mathcal{I}}$, which, by Lemma 1, equals $(n_1 \oplus n_2)^{\mathcal{I}}$; so part 2 is satisfied too, since $n' = n_1 \oplus n_2$. Finally, if the root is modified by the merger, i.e., $n_1$ or $n_2$ is $r$, say $n_1$, then $d = \Upsilon(n_1) = \Upsilon'(n')$, so part 1 of the definition is also satisfied.

Conversely, given arbitrary $d \in G'^{\mathcal{I}}$, let $\Upsilon'$ be the function stipulated by Definition 3 such that $\Upsilon'(r') = d$. Let $\Upsilon$ be the same as $\Upsilon'$ except that $\Upsilon(n_1) = \Upsilon(n')$ and $\Upsilon(n_2) = \Upsilon'(n')$. Then the above argument can be traversed in reverse to verify that $\Upsilon$ satisfies Definition 3, so that $d \in G^{\mathcal{I}}$. ∎

**Lemma 3** *Let $n$ be a node with two mergeable r-edges and let $n'$ be the node with these edges merged. Then $n^{\mathcal{I}} = n'^{\mathcal{I}}$.*

**Proof:** Let the two r-edges be $\langle \mathsf{R}, m_1, M_1, G_1\rangle$ and $\langle \mathsf{R}, m_2, M_2, G_2\rangle$.

Let $d \in n^{\mathcal{I}}$. Then there are between $m_1$ ($m_2$) and $M_1$ ($M_2$) elements of the domain, $d'$, such that $\langle d, d'\rangle \in \mathsf{R}^{\mathcal{I}}$. Therefore there are between the maximum of $m_1$ and $m_2$ and the minimum of $M_1$ and $M_2$ elements of the domain, $d'$, such that $\langle d, d'\rangle \in \mathsf{R}^{\mathcal{I}}$. Also, all $d'$ such that $\langle d, d'\rangle \in \mathsf{R}^{\mathcal{I}}$ are in $G_1^{\mathcal{I}}$ ($G_2^{\mathcal{I}}$). Therefore, all $d'$ such that $\langle d, d'\rangle \in \mathsf{R}^{\mathcal{I}}$ are in $G_1^{\mathcal{I}} \cap G_2^{\mathcal{I}}$, which equals $(G_1 \oplus G_2)^{\mathcal{I}}$ by Lemma 1. Thus $d \in n'^{\mathcal{I}}$.

Let $d \in n'^{\mathcal{I}}$. Then there are between the maximum of $m_1$ and $m_2$ and the minimum of $M_1$ and $M_2$ elements of the domain, $d'$, such that $\langle d, d'\rangle \in \mathsf{R}^{\mathcal{I}}$. Therefore there are between $m_1$ ($m_2$) and $M_1$ ($M_2$) elements of the domain, $d'$, such that $\langle d, d'\rangle \in \mathsf{R}^{\mathcal{I}}$. Also, all $d'$ such that $\langle d, d'\rangle \in \mathsf{R}^{\mathcal{I}}$ are in $(G_1 \oplus G_2)^{\mathcal{I}} = G_1^{\mathcal{I}} \cap G_2^{\mathcal{I}}$. Therefore, all $d'$ such that $\langle d, d'\rangle \in \mathsf{R}^{\mathcal{I}}$ are in $G_1^{\mathcal{I}}$ ($G_2^{\mathcal{I}}$). Therefore $d \in n^{\mathcal{I}}$. ∎

Having dealt with the issue of merging, we can now return to our desired result: showing that "normalization" does not affect the meaning of description graphs.

**Theorem 2** *For all possible worlds $\mathcal{I}$, the extension of the canonical form of a description graph, $G$, resulting from a Basic* CLASSIC *description is the same as the extension of the description.*

**Proof:** Steps 1 and 2 are justified since $G^{\mathcal{I}}$ is a subset of either $\Delta_H$ or $\Delta_C$, which are disjoint.

Step 3 is justified by the fact that, by the definition of description graphs, there must be an element of the domain in the extension of each node in a description graph.

Steps 4, 5, and 6 are easily derived from Definition 3.

Steps 7 and 8 are dealt with in the preceding two lemmas. ∎

## 2.4 Subsumption Algorithm

The final part of the subsumption process is checking to see if a canonical description graph is subsumed by a description. It turns out that it is possible to carry out the subsumption test without the expense of normalizing the candidate subsumer concept.

**Algorithm 1 (Subsumption Algorithm)** *Given a description* D *and description graph* $G = \langle N, E, r\rangle$*,* subsumes?(D, $G$) *is defined to be true if and only if any of the following conditions hold:*

1. *The description graph $G$ is marked incoherent.*

2. D *is equivalent to* THING*. (This is determined by checking first if* D=THING*, or by recursively testing whether* D *subsumes the canonical description graph* $G_{\mathsf{THING}}$*.)*

3. D *is a concept name and is an element of the atoms of $r$.*

4. D *is* $\geq$n R *and some r-edge of $r$ has* R *as its role and min greater than or equal to* n*.*

5. D *is* $\leq$n R *and some r-edge of $r$ has* R *as its role and max less than or equal to* n*.*

6. D *is* $\forall$R:C *and some r-edge of $r$ has* R *as its role and $G'$ as its restriction graph and* subsumes?(C, $G'$)*.*

7. D *is* $\forall$R:C *and subsumes?*(C, $G_{\mathsf{THING}}$) *and $r$ has* CLASSIC-THING *in its atoms.* (Reason: $\forall$R:THING only requires the possibility that R be applicable to an object, which is absent for host values.)

8. D *is* $\forall$A:C *and some a-edge of $G$ is of the form* $\langle r, r', \mathsf{A}\rangle$*, and* subsumes?(C, $\langle N, E, r'\rangle$)*.*

9. D *is* $\forall$A:C *and subsumes?*(C, $G_{\mathsf{THING}}$) *and $r$ has* CLASSIC-THING *in its atoms.*

10. D *is* $\mathsf{A}_1 \circ \ldots \circ \mathsf{A}_n = \mathsf{B}_1 \circ \ldots \circ \mathsf{B}_m$ *and the paths* $\mathsf{A}_1, \ldots, \mathsf{A}_n$ *and* $\mathsf{B}_1, \ldots, \mathsf{B}_m$ *exist in $G$ starting from $r$ and end at the same node.*

11. D *is* $\mathsf{A}_1 \circ \ldots \circ \mathsf{A}_n = \mathsf{B}_1 \circ \ldots \circ \mathsf{B}_m$ *with* $\mathsf{A}_n$ *the same as* $\mathsf{B}_m$ *and the paths* $\mathsf{A}_1, \ldots, \mathsf{A}_{n-1}$ *and* $\mathsf{B}_1, \ldots, \mathsf{B}_{m-1}$ *exist in $G$ starting from $r$ and end at the same node, which has* CLASSIC-THING *in its atoms.* (Reason: If $\mathsf{A}_i^{\mathcal{I}}(\ldots \mathsf{A}_1^{\mathcal{I}}(d)) = \mathsf{B}_j^{\mathcal{I}}(\ldots \mathsf{B}_1^{\mathcal{I}}(d))$ then

$$\mathsf{F}^{\mathcal{I}}(\mathsf{A}_i^{\mathcal{I}}(\ldots \mathsf{A}_1^{\mathcal{I}}(d))) = \mathsf{F}^{\mathcal{I}}(\mathsf{B}_j^{\mathcal{I}}(\ldots \mathsf{B}_1^{\mathcal{I}}(d)))$$

for any attribute F, *as long as the attribute is applicable* (i.e., the value is not in the host domain).)

12. D *is* C $\sqcap$ E *and both subsumes?*(C, $G$) *and subsumes?*(E, $G$) *are true.*

### 2.5 Correctness of Subsumption Algorithm

The soundness of this algorithm is fairly obvious, so we shall not dwell on it. The completeness of the algorithm is, as usual, more difficult to establish. First we have to show that for any canonical description graph or node that is not marked as incoherent, a possible world having a non-empty extension for the description graph or node can be constructed. We will do this in a constructive, inductive manner, constructing a collection of such possible worlds, called the *graphical* worlds of a description graph. A graphical world has a distinguished domain element that is in the extension of the description graph or node.

A common operation is to merge two possible worlds.

**Definition 6** *Let $\mathcal{I}_1$ and $\mathcal{I}_2$ be two possible worlds. The* merge *of $\mathcal{I}_1$ and $\mathcal{I}_2$, $\mathcal{I}_1 \oplus \mathcal{I}_2$, is a possible world with classic realm the disjoint union of the classic realm of $\mathcal{I}_1$ and the classic realm of $\mathcal{I}_2$. The extension of atomic names in $\mathcal{I}_1 \oplus \mathcal{I}_2$ is the disjoint union of their extensions in $\mathcal{I}_1$ and $\mathcal{I}_2$.*

It is easy to show that the extension of a description, a description graph, or a node in $\mathcal{I}_1 \oplus \mathcal{I}_2$ is the union (disjoint union for the classic realm, regular union for the host realm) of its extensions in $\mathcal{I}_1$ and $\mathcal{I}_2$.

Another operation is to add new domain elements to a possible world. These new domain elements must be in the classic realm. The extension of all atomic identifiers remain the same except that the new domain elements belong to some arbitrary set of atomic concept names and have some arbitrary set of fillers (filler) for each role (attribute). Again, it is easy to show that a domain element of the original world is in an extension in the original world iff it is in the extension in the augmented world.

Given a node, $n$, that is not marked as incoherent, we construct the graphical worlds for $n$ as follows:

1. If the atoms of $n$ are precisely THING, then $n$ can have no r-edges, because the only constructs that cause r-edges to be created also add CLASSIC-THING to the atoms. Any possible world, with any domain element the distinguished domain element, is a graphical world for $n$.

2. If the atoms of $n$ include HOST-THING, then $n$ can have no r-edges. Any possible world, with distinguished element any domain element in the extension of all the atoms of $n$ and in no other host concepts, is a graphical world for $n$. (Because of the requirements on the host domain, there are an infinite number of these domain elements.)

3. If the atoms of $n$ include CLASSIC-THING, then for each r-edge, $\langle \mathsf{R}, m, M, G\rangle$, in $n$, construct between $m$ and $M$ graphical worlds for $G$. This can be done for any number between $m$ and $M$ because if $m > 0$ then $G$ is not marked incoherent, and if $G$ is marked incoherent then $M = 0$.

   No two of these graphical worlds should have the same host domain element as their distinguished element. (Again, this is possible because the extension of a host concept is either empty or infinite.) Now merge all the graphical worlds for each r-edge into one possible world. Add some new domain elements such that one of them is in exactly the extensions of the atoms of $n$ and has as fillers for each R exactly the distinguished elements of the appropriate graphical worlds. This domain element will have the correct number of fillers for each r-edge, because of the disjoint union of the classic realms in the merge process and because of the different host domain elements picked above; therefore it is in the extension of $n$. Thus the resulting world is a graphical world for $n$.

Given a description graph, $G = \langle N, E, r\rangle$, that is not marked incoherent, we construct the graphical worlds for $G$ as follows: For each node $n \in N$ construct a graphical world for $n$. This can be done because none of them are marked incoherent. Merge these graphical worlds. Modify the resulting world so that for each $\langle n_1, n_2, \mathsf{A}\rangle \in E$ the A-filler for the distinguished node of the graphical world from $n_1$ is the distinguished node of the graphical world from $n_2$. It is easy to show that the distinguished node of the graphical world of $r$ is in the extension of $G$, making this a graphical world for $G$.

Now we can show the final part of the result.

**Theorem 3** *If the subsumption algorithm indicates that the canonical description of some graph $G$ is not subsumed by the Basic* CLASSIC *description* D*, then for some possible world there is a domain element in the extension of the graph but not in the extension of* D*. Therefore $G$ is not subsumed by* D*.*

**Proof:** The proof actually shows that if the subsumption algorithm indicates that some canonical description graph, $G$, is not subsumed by some description, D, then there are some graphical worlds for $G$ such that their distinguished domain elements are not in the extension of D. Remember that the subsumption algorithm indicates that $G$ is not subsumed by D, so $G$ must not be marked as incoherent and thus there are graphical worlds for $G$.

The proof proceeds by structural induction on D. Let $G = \langle N, E, r\rangle$.

- If D is an atomic concept name or a pre-defined host concept, then D does not occur in the atoms of $r$. By construction, in any graphical world for $G$ the distinguished domain element will not be in the extension of D. Similarly, if D is CLASSIC-THING or HOST-THING, then the distinguished domain elements will be in the wrong realm. If D is THING, then it is not possible for the subsumption algorithm to indicate a non-subsumption. In each case any graphical world for $G$ has the property that its distinguished domain element is not in the extension of D.

- If D is of the form $\mathsf{D}_1 \sqcap \mathsf{D}_2$ then the subsumption algorithm must indicate that $G$ is not subsumed by at least one of $\mathsf{D}_1$ or $\mathsf{D}_2$. By the inductive hypothesis, we get some graphical worlds of $G$ where the distinguished domain elements are not in the extension of $\mathsf{D}_1$ or not in the extension of $\mathsf{D}_2$, and thus are not in the extension of D.

- If D is the form $\geq$n R then either the r-edge from $r$ labelled with R has min less than n or there is no such r-edge.

  In the former case there are graphical worlds for $G$ in which the distinguished node has $\mathsf{n} - 1$ fillers for R, because n is greater than the min on the r-edge for R, and thus the distinguished node is not in the extension of D.

In the latter case, there are graphical worlds for $G$ in which its distinguished node has any number of fillers for R. Those with $n - 1$ fillers have the property that their distinguished node is not in the extension of D.

- If D is of the form $\leq$n R then either the r-edge from $r$ labelled with R has max greater than n (including $\infty$) or there is no such r-edge.

  In the former case there are graphical worlds for $G$ in which the distinguished node has $n + 1$ fillers for R, because n is less than the max on the r-edge for R, and thus the distinguished node is not in the extension of D.

  In the latter case, there are graphical worlds for $G$ in which its distinguished node has any number of fillers for R. Those with $n + 1$ fillers have the property that their distinguished node is not in the extension of D.

- If D is of the form $\forall$R:C, where R is a role, then two cases arise.

  1. If subsumes?(C, $G_{\mathsf{THING}}$) then CLASSIC-THING is not in the atoms of $r$. Then there are some graphical worlds for $G$ whose distinguished element is in the host realm, and thus not in the extension of D.
  2. Otherwise, either there is an r-edge from $r$ with role R and description graph $H$ such that subsumes?(C, $H$) is false or there is no r-edge from $r$ with role R. Note that the extension of C is not the entire domain, and thus must be a subset of either the host realm or the classic realm.

     In the former case $H$ is not marked incoherent (or else the subsumption could not be false) and the max on the r-edge cannot be 0. Thus there are graphical worlds for $H$ whose distinguished element is not in the extension of C and there are graphical worlds for $G$ that use these graphical worlds for $H$ as distinguished domain element R-fillers. In these graphical worlds for $G$ the distinguished element is not in the extension of D.

     In the latter case, pick graphical worlds for $G$ that have some distinguished node R-filler in the wrong realm. In these graphical worlds for $G$ the distinguished element is not in the extension of D.

- If D is of the form $\forall$A:C where A is an attribute then two cases arise.

  1. If subsumes?(C, $G_{\mathsf{THING}}$) then CLASSIC-THING is not in the atoms of $r$. Then there are some graphical worlds for $G$ whose distinguished element is in the host realm, and thus not in the extension of D.
  2. Otherwise, either there is an a-edge from $r$ with attribute A to some other node $r'$ such that subsumes?(C, $H$) is false, where $H = \langle N, E, r' \rangle$; or there is no a-edge from $r$ with attribute A. Note that the extension of C is not the entire domain, and thus must be a subset of either the host realm or the classic realm.

     In the former case $H$ is not marked incoherent, because $G$ is not marked incoherent. Thus there are graphical worlds for $H$ whose distinguished element is not in the extension of C. Given any graphical world for $H$, a graphical world for $G$ can be formed simply by changing the distinguished domain element. If

the original graphical world's distinguished element is not in the extension of C, then the new graphical world's distinguished element will not be in the extension of D, as required.

In the latter case, pick graphical worlds for $G$ that have their distinguished node A-filler in the wrong realm. In these graphical worlds for $G$ the distinguished element is not in the extension of D.

- If D is of the form $A_1 \circ \ldots \circ A_n = B_1 \circ \ldots \circ B_m$ several cases again arise.

  1. If one of the paths $A_1, \ldots, A_{n-1}$ or $B_1, \ldots, B_{m-1}$ does not exist in $G$ starting from $r$, then find the end of the partial path and use graphical worlds in which the domain element for this node has an element of the host domain as its filler for the next attribute in the path. Then one of the full paths will have no filler.
  2. If the paths $A_1, \ldots, A_n$ and $B_1, \ldots, B_m$ exist in $G$ starting from $r$ but end at different nodes, then use graphical worlds in which the domain elements for these two nodes are different.
  3. If one of the paths $A_1, \ldots, A_n$ and $B_1, \ldots, B_m$ does not exist in $G$ starting from $r$ but the paths $A_1, \ldots, A_{n-1}$ and $B_1, \ldots, B_{m-1}$ both exist in $G$ starting from $r$ and end at the same node then either CLASSIC-THING is not in the atoms of this node or $A_n \neq B_m$. In the former case use graphical worlds in which the domain element for this node is in the host realm. In the latter case use graphical worlds that have different fillers for $A_n$ and $B_m$ for the domain element for this node.
  4. If one of the paths $A_1, \ldots, A_n$ and $B_1, \ldots, B_m$ does not exist in $G$ starting from $r$ but the paths $A_1, \ldots, A_{n-1}$ and $B_1, \ldots, B_{m-1}$ both exist in $G$ starting from $r$ and end at different nodes then use graphical worlds that have different fillers for the domain elements of these nodes or that have the domain elements in the host realm.

  In all cases we have that either one of ${A_n}^{\mathcal{I}}(\ldots {A_1}^{\mathcal{I}})(d)$ or ${B_m}^{\mathcal{I}}(\ldots {B_1}^{\mathcal{I}})(d)$ does not exist or ${A_n}^{\mathcal{I}}(\ldots {A_1}^{\mathcal{I}})(d) \neq {B_m}^{\mathcal{I}}(\ldots {B_1}^{\mathcal{I}})(d)$, so the distinguished domain element is not in the extension of D.

### 2.6 Implementing the subsumption algorithm

In this section we provide some further comments about the actual subsumption algorithm used by the CLASSIC system, including a rough analysis of its complexity.

As we have described it, deciding whether description C subsumes D is accomplished in three phases:

1. Convert D into a description graph $G_D$.
2. Normalize $G_D$.
3. Verify whether C subsumes $G_D$.

**Step 1**: Conversion is accomplished by a simple recursive descent parser, which takes advantage of the fact that the syntax of description logics (i.e., the leading term constructor) makes them amenable to predictive parsing. Clearly, constructing graphs for fixed sized terms (like `at-least`) is constant time (if we measure size so that an integer is size 1 no matter how large), while the time for non-recursive terms (like `same-as`) is proportional to their length. Finally, recursive terms (like `all`, `and`) only require a fixed amount of additional work, on top of the recursive processing. Therefore, the first stage can be accomplished in time proportional to the size of the input description. In order to speed up later processing, it will be useful to maintain various lists, such as the lists of atomic concept identifiers, or roles/attributes, in sorted order. This sorting needs to be done initially (later, ordering will be maintained by performing list merges) and this incurs, in the worst case a quadratic overhead in processing[8]. In any case, the total size of the graph constructed (including the sizes of the nodes, etc.) is proportional to the size of the original concept description.

**Step 3:** Checking whether a description $C$ subsumes a description graph $G_D$, can be seen to run in time proportional to the size of the subsuming concept, modulo the cost of lookups in various lists. Since these are sorted, the lookup costs are bounded by the logarithm of the size of the candidate subsumee graph, so the total cost is bounded by $O(\mid C \mid \; * \; log \mid G_D \mid)$.

**Step 2**: Normalization is accomplished by a post-order traversal of the description graph: in processing a description *graph* $\langle N, E, r\rangle$, each node in $N$ is normalized first independently (see details below), and afterwards the attribute edges $E$ are normalized. This later task involves identifying multiple identically-labelled attribute edges leaving a node (this is done in one pass since the attribute edges are grouped by source node, and sorted by attribute name), and "merging" them. Merging two edges is quite easy in and of itself, but when merging the nodes at their tips, we must be careful because node mergers may cascade; for example, if a concept has the form $\mathsf{a_1} = \mathsf{b_1} \sqcap \mathsf{a_2} = \mathsf{b_2} \sqcap \ldots \sqcap \mathsf{a_n} = \mathsf{b_n} \sqcap \mathsf{a_1} = \mathsf{a_2} \sqcap \mathsf{a_2} = \mathsf{a_3} \sqcap \ldots \sqcap \mathsf{a_{n-1}} = \mathsf{a_n}$ then the original graph will have $2n+1$ nodes, but $2n$ of these are collapsed by normalization step 8. To discover this efficiently, we use a version of A¨-Kaci's algorithm for unifying $\Psi$-terms (Aït-Kaci, 1984; Aït-Kaci & Nasr, 1986); the algorithm relies on the UNION-FIND technique to identify nodes to be merged, and runs in time just slightly more than linear in the number of nodes in $N$. Therefore the cost of the non-recursive portion of graph normalization is roughly linear in the number of nodes in it.

The merging of two description graph *nodes* is quite similar to the normalization of a single node: the atomic concept identifier lists need to sorted/merged, with duplicates eliminated on the fly. This can be done in time proportional to the *size of the nodes* themselves, if we make the size of the node include the size of the various lists in it, such as atoms. The processing of role edges leaving a node is, again, one of identifying and merging identically-labelled edges. (But in this case the mergers of labelled edges do not interact, so a single pass over the role-edge list is sufficient.) The cost of non-recursive aspects of any such merger is once again proportional to the size of the local information.

We are therefore left with the problem of bounding the *total number* of procedure calls to NormalizeGraph, NormalizeNode, MergeEdge, and MergeNode, and then bounding the sizes of the nodes being merged.

NormalizeGraph and NormalizeNode are called exactly once on every (sub)graph and node in the original graph, as part of the depth-first traversal, and as argued above, on

8. We tend not to use fancy sorting techniques since these lists are not likely to be very long.

their own they contribute at most time proportional to the total size of the original graph, which was proportional to the size of the original description.

The number of calls to MergeEdge and MergeNode is not so simply bounded however – the same node may be merged several times with others. However, these calls are paired, and each invocation of MergeNode reduces the number of nodes in the graph by one. Therefore, since the number of nodes is not incremented elsewhere, the total number of calls to MergeEdge and MergeNode is bounded by the number of nodes in the original graph. The only problem is that the non-recursive cost of a call to MergeNode depends on the size of the argument nodes, and each call may increase the size of the remaining node to be the sum of the sizes of the two original nodes.

Therefore, if the original concept had size S, with the graph having $n$ nodes, each of size $v_i$, then the worst case cost would result from the iterative summation of sizes:

$$\begin{aligned}&(v_{i_1} + v_{i_2}) + (v_{i_1} + v_{i_2} + v_{i_3}) + (v_{i_1} + v_{i_2} + v_{i_3} + v_{i_4}) + \ldots \\ &= n * v_{i_1} + (n-1) * v_{i_2} + \ldots + 1 * v_{i_n}\end{aligned}$$

Given that $n$ and all $v_j$ are bounded by $S$, clearly the above is in the worst case $O(S^3)$. In fact, given the constraint that $\sum_{j=1} nv_j = S$, it is possible to argue that the worst case cost will occur when $v_j = 1$ for every j, (i.e., when $n = S$), in which case the cost is really just $O(S^2)$.

There are other theoretical improvements that could be attempted for the algorithm (e.g., merging nodes in the correct order of increasing size) as well as its analysis (e.g., only nodes in graphs at the same depth in the tree can be merged).

We remark that like all other description logics, CLASSIC permits identifiers to be associated with complex descriptions and then these identifiers can be used in other descriptions (though no recursion is allowed). The expansion of identifiers is a standard operation which can lead to exponential growth in size in certain pathological cases (Nebel, 1990), making the subsumption problem inherently intractable. As with the type system of the programming language Standard ML, such pathological cases are not encountered in practice, and the correct algorithm is simple, straightforward and efficient in normal cases (unlike the correct algorithm for reasoning with the set constructor, say).

Because users rarely ask only whether some concept subsumes another, but rather are interested in the relationship between pairs of concepts, CLASSIC in fact constructs the normalized description graph of any description given to it. This suggests that it might be better to check whether one description graph subsumes another one, rather than checking whether a description subsumes a graph. In general, this works quite well, except that we would have to verify that the attribute edges in the subsumer graph form a *subgraph* of the subsumee's attribute edges. Since edges are uniquely labelled after normalization, this is not inherently hard, but it still requires a complete traversal (and hence marking/unmarking) of the upper graph. We have therefore found it useful to encode as part of the description graph's root the `same-as` restrictions that lead to the construction of the corresponding a-edges; then, during subsumption testing, the only aspect of the subsumer related to `same-as` which is checked is this list of `same-as` pairs.

Also, the above description of the algorithm has tried to optimize the cost of normalization, which dominates when checking a single subsumption. If in the overall use of a system (e.g., processing individuals), inquiries about the restrictions on roles/attributes are frequent, and space usage is not a problem, then it may be practically advantageous to maintain the r-edges and a-edges of a node in a hash table, rather than a sorted list, in order to speed up access. (Note that for merging r-edges, one must however still have some way of iterating through all the values stored in the hash table.)

## 3. Individuals in Descriptions

In practical applications where DLs have been used, such as integrity constraint checking, it is often very useful to be able to specify ranges of atomic values for roles. The most common examples of this involve integers, e.g., *"the year of a student can be 1,2,3 or 4"*, or what are called enumerated types in Pascal, e.g., *"the gender of a person is either M or F"*. One way to allow such constraints is to introduce a new description constructor, a set description, which creates a description from a list of individual names, and whose obvious extension is the set consisting of the extensions of the individuals that appear in the list. This construct could be used in terms like ∀year:{1 2 3 4}. Another useful constructor involving individuals is a fills restriction, p : I, which denotes objects that have the extension of the individual I as one of the fillers for the relationship denoted by role or attribute p. (Note that for an attribute, q, ∀q:{I} is the same as q : I.)

Within the paradigm of DLs, these constructors are quite useful and can in fact be used to express new forms of incomplete information. For example, if we only know that Ringo is in his early fifties, we can simply assert that Ringo is described by ∀age:{50 51 52 53 54}. The constructors can also be used to ask very useful queries. For example, to find all the male persons it suffices to determine the instances of gender : M.

The new constructors do interact with previous ones, such as cardinality constraints: clearly the size of a set is an upper cardinality bound for any role it restricts. This interaction is not problematic as long as the individuals in the set are *host values*, since such individuals have properties that are fixed and known ahead of time. However, once we allow classic individuals as members of sets, then the properties of these individuals might themselves affect subsumption. As a simple example, if we know that Ringo is an instance of the concept ROCK-SINGER (which we shall write as Ringo ∈ ROCK-SINGER) then the extension of ∀friends:ROCK-SINGER is always a superset of the extension of ∀friends:{Ringo}.

This is disturbing because then the classification hierarchy of definitions would change as new facts about individuals are added to the knowledge base. Definitions are not meant to be contingent of facts about the current world. Therefore, subsumption is usually defined to be independent of these "contingent" assertions. As we shall see below, the use of individual properties in description subsumption also leads to intractability.

### 3.1 Complex Subsumption Reasoning: An Example

Traditional proofs of intractability (e.g. (Levesque & Brachman, 1987)) have occasionally left users of DLs puzzled over the intuitive aspects of a language which make reasoning difficult. For this reason we present an example that illustrates the complexity of reasoning with the set description.

Suppose that we have the concept of JADED-PERSON as being one who wants only to visit the Arctic and/or the Antarctic, wherever there are penguins:

$$\mathsf{JADED\text{-}PERSON} \doteq \forall \mathsf{wantsToVisit}{:}(\{\mathsf{Arctic\ Antarctic}\} \sqcap \forall \mathsf{hasPenguins!}{:}\{\mathsf{Yes}\})$$

Suppose we do not remember which is the Arctic and which the Antarctic; but we do know that the South Pole is located in one of these two places, and that there are penguins there, while the North Pole is located in one of these two places, and there are no penguins there. Assuming that isLocatedIn! and hasPenguins! are attributes—roles with exactly one filler, we can record

$$\mathsf{Southpole} \in \forall \mathsf{isLocatedIn!}{:}(\{\mathsf{Arctic\ Antarctic}\} \sqcap \forall \mathsf{hasPenguins!}{:}\{\mathsf{Yes}\})$$
$$\mathsf{Northpole} \in \forall \mathsf{isLocatedIn!}{:}(\{\mathsf{Arctic\ Antarctic}\} \sqcap \forall \mathsf{hasPenguins!}{:}\{\mathsf{No}\})$$

We are thus unable to distinguish the exact location of the Southpole and Northpole; however, since hasPenguins! has a single filler, exactly one of Arctic and Antarctic can (and in fact must) have Yes as filler for hasPenguins!, and therefore exactly one of them is the location of Southpole.

As a result of these facts, we know that the extension of JADED-PERSON must be a subset of the extension of $\leq$1 wantsToVisit in any database containing the above facts about Southpole and Northpole.

Observe that we have here not just an occasional worse-case behavior, but a generalized difficulty in reasoning with set descriptions. Because subsumption ignores assertions about individuals, this does not (yet) show that subsumption per se must perform these inferences. A simple transformation, given in the appendix, establishes this fact, by converting the recognition of individuals into a question about the subsumption of two descriptions by making all the individuals involved attribute-fillers for new dummy attributes, and their descriptions as restrictions on these attributes. As a result, if the description is non-empty then these attribute values must satisfy the corresponding restrictions.

### 3.2 A Modified Semantics for Individuals

We have seen two problems with individuals appearing in descriptions: (1) the effect of "mutable facts" on extensional relationships between "immutable" descriptions, and (2) the computational intractability of subsumption caused by the appearance of individuals in descriptions.

To deal with the first problem, it is reasonable to restrict the computation of subsumption so that it cannot access "database facts" about individuals, such as their role fillers, so that all individuals are treated like host identifiers. This is a procedural description of some aspect of reasoning, in the same sense as negation-by-failure is in Prolog. As with Prolog, it would be desirable to find a semantic account of this phenomenon.

A semantics that ignores mutable facts when determining subsumption is not hard to devise—all that is required is to have two different sets of possible worlds corresponding to a KB containing both concepts and individuals. One set consists of all possible worlds that model all the information in the KB; the second consists of all possible worlds that model only the information about concepts (and roles and attributes). When asking questions about individuals, the first set of possible worlds must be considered; when asking subsumption questions, the second, larger, set must be considered, thus ignoring any effects of the mutable facts.

However, this semantics does not solve the computational problem with individuals in descriptions. To deal with this problem, the semantics of individuals are modified as follows: instead of mapping individuals into separate elements of the domain, as is done in a standard semantics, individuals are mapped into *disjoint subsets of the domain*, intuitively representing different possible realizations of that (Platonic) individual.

Therefore, the semantics of the set constructor is now stated as follows: Domain value $d$ belongs to the extension of $\{\mathsf{B}_1 \ldots \mathsf{B}_n\}$ iff $d$ belongs to the extension of one of the $\mathsf{B}_i$. An associated change in the notion of cardinality is required—two elements of the domain are considered *congruent* if they belong to the extension of the same individual or if they are identical. The cardinality of a set of elements of the domain is then the size of the set modulo this congruence relationship. This means that occurrences of different identifiers in description(s) are guaranteed to be unequal, but distinct occurrences of the same individual identifier are not guaranteed to denote the same individual.

Here are two consequences of this stance:

1. Looking at the descriptions of Southpole and Northpole in Section 3.1, the distinct occurrences of Arctic might be satisfied by distinct domain elements, with different role fillers. (In greater detail: the extension of Arctic might include domain elements $d_1$ and $d_2$, with $d_1$ satisfying condition hasPenguins! : Yes, while $d_2$ satisfies hasPenguins! : No. If Southpole is then located in $d_1$, while Northpole is located in $d_2$, then we still have both satisfying isLocatedIn! : Arctic. Similarly for domain elements $d_3$ and $d_4$ in the extension of Antarctic. Therefore one could have two places to visit where there are penguins, $d_1$ and $d_3$.)

2. Even though an individual may have a description that includes

   isLocatedIn! : Arctic $\sqcap$ originatesIn! : Arctic,

   it need not satisfy the condition isLocatedIn! = originatesIn!, since the equality restriction requires identity of domain values.

## 4. Adding Individuals to CLASSIC

Individuals can occur in both classic and host descriptions. The following constructs create classic descriptions:

R : I
A : I
$\{\mathsf{I}_1 \ldots \mathsf{I}_n\}$

where A is an attribute, R is a role, I is the name of a classic individual or a host value, collectively called individuals, and $\mathsf{I}_j$ are names of classic individuals. New host descriptions can be constructed using $\{\mathsf{I}_1 \ldots \mathsf{I}_n\}$, where the $\mathsf{I}_j$ are host values.

The interpretation function $\cdot^{\mathcal{I}}$ is extended to individual identifiers, by requiring that $\mathsf{I}^{\mathcal{I}}$ be a non-empty subset of $\Delta_C$, if I is syntactically not recognized to be a host individual, and making $\mathsf{I}^{\mathcal{I}} = \{\mathsf{I}\}$ for host values I. As stated earlier, the interpretations of distinct identifiers must be non-overlapping.

The interpretation $\mathsf{C}^{\mathcal{I}}$ of non-atomic descriptions is modified as follows:

- $\mathsf{p}:\mathsf{I}^{\mathcal{I}} = \{d \in \Delta_C \mid \exists x\ (d,x) \in \mathsf{p}^{\mathcal{I}}\ \wedge\ x \in \mathsf{I}^{\mathcal{I}}\}$

- $\{\mathsf{I}_1 \ldots \mathsf{I}_n\}^{\mathcal{I}} = \bigcup_k \mathsf{I}_k^{\mathcal{I}}$ if the $\mathsf{I}_k$ are all classic individuals; $\{\mathsf{I}_1 \ldots \mathsf{I}_n\}^{\mathcal{I}} = \{\mathsf{I}_1 \ldots \mathsf{I}_n\}$ if $\mathsf{I}_k$ are all host individuals; empty otherwise.

- $(\geq \mathsf{n}\ \mathsf{p})^{\mathcal{I}}$ (resp. $(\leq \mathsf{n}\ \mathsf{p})^{\mathcal{I}}$) is those objects in $\Delta_C$ with at least (resp. at most) n *non-congruent* fillers for role p

The development of the subsumption algorithm in Section 2 is then modified to take into account the added constructs with the modified semantics introduced earlier.

First description graphs are extended. A node of a description graph is given a third field, which is either a finite set of individuals or a special marker denoting the "universal" set. This field is often called the *dom* of the node. Both a-edges and r-edges are given an extra field, called the *fillers* of the edge. This field is a finite set of individuals. Where unspecified, as in constructions in previous sections, the *dom* of a node is the universal set and the *fillers* of an a-edge or an r-edge is the empty set.

The semantics of description graphs in Definition 3 are extended to the following:

**Definition 7** *Let $G = \langle N, E, r\rangle$ be a description graph and let $\mathcal{I}$ be a possible world.*

*An element, $d$, of $\Delta$ is in $G^{\mathcal{I}}$, iff there is some function, $\Upsilon$, from $N$ into $\Delta$ such that*

1. *$d = \Upsilon(r)$;*
2. *for all $n \in N$ $\Upsilon(n) \in n^{\mathcal{I}}$;*
3. *for all $\langle n_1, n_2, \mathsf{A}, F\rangle \in E$ we have $\langle \Upsilon(n_1), \Upsilon(n_2)\rangle \in \mathsf{A}^{\mathcal{I}}$, and for all $\mathsf{f} \in F$, $\Upsilon(n_2) \in \mathsf{f}^{\mathcal{I}}$.*

*An element, $d$, of $\Delta$ is in $n^{\mathcal{I}}$, where $n = \langle C, H, S\rangle$, iff*

1. *for all $\mathsf{C} \in C$, we have $d \in \mathsf{C}^{\mathcal{I}}$;*
2. *for all $\langle \mathsf{R}, m, M, G, F\rangle \in H$,*
   (a) *there are between $m$ and $M$ elements, $d'$, of the domain such that $\langle d, d'\rangle \in \mathsf{R}^{\mathcal{I}}$;*
   (b) *$d' \in G^{\mathcal{I}}$ for all $d'$ such that $\langle d, d'\rangle \in \mathsf{R}^{\mathcal{I}}$; and*
   (c) *for all $\mathsf{f} \in F$ there is a domain element, $d'$, such that $\langle d, d'\rangle \in \mathsf{R}^{\mathcal{I}}$ and $d' \in \mathsf{f}^{\mathcal{I}}$*
3. *If the $S$ is not the universal set then $\exists \mathsf{f} \in S$ such that $d \in \mathsf{f}^{\mathcal{I}}$.*

When merging nodes, the *dom* sets are intersected. Merging description graphs is unchanged. When merging a-edges and r-edges, the sets of fillers are unioned.

The translation of descriptions into description graphs is extended by the following rules:

8. A description of the form R:I is turned into a description graph with one node and no a-edges. The node has as its atoms CLASSIC-THING and a single r-edge with role R, min 0, max $\infty$, and fillers {I}. The description graph restricting this r-edge is $G_{\mathsf{CLASSIC\text{-}THING}}$ if I is a classic individual, and $G_{\mathsf{HOST\text{-}THING}}$ otherwise.

9. A description of the form A : I is turned into a description graph with two nodes with a single a-edge between them. The distinguished node of the graph is the source of the a-edge. It has no r-edges and has as atoms CLASSIC-THING. The other node also has no r-edges. It has as atoms CLASSIC-THING if I is a classic individual, and HOST-THING otherwise. The a-edge has as its single filler I.

10. A description of the form $\{I_1 \ldots I_n\}$ is turned into a description graph with one node. The node has as *dom* the set containing $I_1$ through $I_n$, and no r-edges. The atoms of the node is HOST-THING if all of the individuals are host values, and CLASSIC-THING if all of the individuals are classic individual names. (Note that the parser ensures that individuals either must all be host values or must all be classic individual names.)

A short examination shows that Theorem 1 is true for these graphs, i.e., the extension of description graphs formed using these rules is the same as the extension of the description from which they were formed.

The following transformations are added to the canonicalization algorithm:

9. If the *dom* of a node is empty, mark the node incoherent.

10. If a host value in the *dom* of a node is not in all the atoms of the node, remove it from the *dom*.

11. If an a-edge has more than one filler, then mark the description graph as incoherent.

12. If an a-edge has a filler and the node at its end has the universal *dom*, make the *dom* be the filler.

13. If the filler of an a-edge is not included in the *dom* of the node at its end, mark the description graph as incoherent.

14. If a node has only one element in its *dom*, make this element be the filler for all the a-edges pointing to it.

15. If the fillers of some r-edge are not a subset of the *dom* of the distinguished node of the restriction graph of the edge, mark the node of the r-edge incoherent.

16. If the min on an r-edge is less than the cardinality of fillers on it, let the min be this cardinality.

17. If the max on an r-edge is greater than the cardinality of the *dom* on the distinguished node of the description graph of the r-edge, make the max of this edge be the cardinality of the *dom*.

18. If the min on an r-edge is greater than or equal to the cardinality of the *dom* on the distinguished node of the restriction graph of the r-edge, let the fillers of the edge be the union of its fillers and the *dom* above. (If min is greater than the cardinality, then steps 4 and 17 detect the inconsistency.)

19. If the max on an edge is equal to the cardinality of fillers on the edge, let the *dom* on the distinguished node of the description graph of the r-edge be the intersection of the *dom* and the fillers. (If max is less than the cardinality, steps 18 and 4 detect the inconsistency.)

Note that in the new canonical form all a-edges pointing to a single node have the same value for their fillers, and that if this is not the empty set, then the node has this set as the value for its *dom*.

The proofs of Lemmas 3 and 2 also work for this extension of description graphs. The proof of Theorem 2 can then be extended for these graphs.

The subsumption algorithm from page 289 is extended as follows:

13. D is R : I and some r-edge of $r$ has role R and fillers including I.

14. D is A : I and some a-edge from $r$ has attribute A and fillers including I.

15. D is $\{I_1 \ldots I_n\}$ and the *dom* of $r$ is a subset of $\{I_1 \ldots I_n\}$.

Again, the soundness of the extended algorithm is fairly obvious. The completeness proof has the following additions to the construction of graphical worlds:

- The extension of classic individual names starts out empty.
- When constructing graphical worlds for a node that includes HOST-THING in its atoms and has a non-universal *dom*, pick only those domain elements corresponding to the elements of its *dom*.
- When constructing graphical worlds for a node that includes CLASSIC-THING in its atoms and has a non-universal *dom*, add the distinguished domain element to the extension of one of its *dom* elements.
- When constructing graphical worlds for the r-edges of a node, ensure that each element of the fillers of the r-edge has the distinguished element of at least one of the graphical worlds in its extension by either adding them to the extension or using appropriate host domain elements. (This can be done because the fillers must be a subset of the *dom* of the distinguished node of the graphical world and any host values must belong to its atoms.)

The fillers for a-edges need not be considered here because they are "pushed" onto the nodes in the canonicalization process.

The proof of Theorem 3 is then extended with the following cases:

- If D is of the form $\{I_1 \ldots I_n\}$ then the *dom* of $r$ is not a subset of $\{I_1, \ldots, I_n\}$. Thus there are graphical worlds for $G$ in which the distinguished domain element is not in the extension of any of the $I_j$.
- If D if of the form A : I then either the a-edge from $r$ labelled with A does not have filler I or there is no such a-edge.

In the former case the node pointed to by the a-edge cannot have as its domain the singleton consisting of I. Therefore there are graphical worlds for $G$ that have their distinguished node A-filler not in the extension of I, as required.

In the latter case, pick graphical worlds for $G$ that have their distinguished node A-filler in the wrong realm. In these graphical worlds for $G$ the distinguished element is not in the extension of D.

- If D is of the form R : I then either the r-edge from $r$ labelled with R does not have filler I or there is no such r-edge.

  In the former case either the cardinality of the *dom* of the distinguished node of the description graph of this r-edge is greater than the min, m, of the r-edge, or the *dom* does not include I. If the *dom* does not include I, then all graphical worlds for the node have their distinguished element not in the extension of I, as required. If the *dom* does include I, then there are at least m elements of the *dom* besides I, and the fillers of the r-edge are a subset of the set of these elements. There are thus graphical worlds for $G$ that use only these elements, as required.

  In the latter case, pick graphical worlds for $G$ that have some distinguished node R-filler in the wrong realm. In these graphical worlds for $G$ the distinguished element is not in the extension of D.

This shows that the subsumption algorithm given here is sound and complete for the modified semantics presented here.

## 5. Complete CLASSIC

We now make a final pass to deal with some less problematic aspects of CLASSIC descriptions that have not been appropriately covered so far.

CLASSIC allows primitive descriptions of the form (**PRIMITIVE** D T), where D is a description, and T is a symbol. The extension of this is some arbitrary subset of the extension of D, but is the same as the extension of (**PRIMITIVE** E T), provided that D and E subsume each other. In this way one can express EMPLOYEE, a kind of a person who must have an employee number, as

(**PRIMITIVE** (PERSON ⊓ ≥1 employeeNr) employee)

This construct can be removed by creating for every such primitive an atomic concept (e.g., EMPLOYEEHOOD) and then replacing the definition of the concept by the conjunction of the necessary conditions and this atom, in this case EMPLOYEEHOOD ⊓ (PERSON ⊓ ≥1 employeeNr). Care has to be taken to use the same atomic concept for equivalent primitives.

CLASSIC permits the declaration of disjoint primitives, essentially allowing one to state that the extensions of various atomic concepts must be disjoint in all possible worlds. To deal with such declarations, we need only modify the algorithm for creating canonical graphs by adding a step that marks a node as incoherent whenever its atoms contains two identifiers that have been declared to be disjoint.

To allow an approximate representation for ideas that cannot be encoded using the constructors expressly provided, CLASSIC allows the use of *test-defined concepts*, using the following syntax:

(**TEST** *[host-language Boolean function]*)

e.g., (**TEST** Prime-Number-Testing-Function).[9] For the purposes of subsumption, these are treated as "black-boxes", with semantics assigned as for atomic concepts. (Test concepts have a real effect on reasoning at the level of individuals, where they can perform constraint checking.)

With these simple additions, the above algorithm is a sound and complete subsumption algorithm for descriptions in CLASSIC 1, under the modified semantics introduced in this paper.

## 6. Summary, Related Work, and Conclusions

We believe this paper makes two kinds of contributions: First, the paper presents an abstracted form of the subsumption algorithm for the CLASSIC description logic, and shows that it is efficient and correct under the modified semantics. This is significant because previous claims of correct and efficient subsumption algorithms in implemented DLs such as KANDOR (Patel-Schneider, 1984) and CANDIDE (Beck et al., 1989) have turned out to be unfounded (Nebel, 1988).

A tractability proof for a language like Basic CLASSIC is claimed to exist (but is not proven) in (Donini et al., 1991), and an alternate proof technique may be found by considering a restriction of the (corrected) subsumption algorithm in (Hollunder & Nutt, 1990).

Description graphs have also turned out to be of interest because they support further theoretical results about DLs, concerning their learnability (Cohen & Hirsh, 1994; Pitt & Frazier, 1994)—results which would seem harder to obtain using the standard notation for DLs.

Second, this paper investigates the effect of allowing individuals to appear in descriptions of DLs. As independently demonstrated in (Lenzerini & Schaerf, 1991), adding a set description introduces yet another source of intractability, and we have provided an intuitive example illustrating the source of difficulties. The implementers of the CLASSIC system, like others who do not use refutation/tableaux theorem-proving techniques, chose not to perform all inferences validated by a standard semantics, not just because of the formal intractability result but because no obvious algorithm was apparent, short of enumerating all possible ways of filling roles. The subset of inferences actually performed was initially described procedurally: "facts" about individuals were not taken into account in the subsumption algorithm. This paper provides a denotational semantic account for this incomplete set of inferences. The formal proof of this being a correct account is a corollary of the completeness proof for the subsumption algorithm in Section 4, and the observation that the graph construction and subsumption algorithms in that section do indeed ignore

9. In order to deal with the two realms, CLASSIC in fact provides two constructors: **H-TEST** and **C-TEST**, for host and classic descriptions, but this does not cause any added complications besides keeping track of the correct realm.

the properties of the individuals involved. The one difference between the original implementation of CLASSIC and the current semantics is that attribute paths ending with the same filler were used to imply an equality condition. As noted in Section 3.2, the modified semantics does not support this inference, and it was taken out of the implementation of CLASSIC. It is significant that the change to the standard semantics is small, easy to explain to users (either procedurally or semantically), and only affects the desired aspects of the language (i.e., all reasoning with Basic CLASSIC remains exactly as before).

### Acknowledgments

We wish to thank Ronald Brachman and our other colleagues in the CLASSIC project for their collaboration, and the JAIR referees for their excellent suggestions for improving the paper. In particular, one of the referees deserves a medal for the thoroughness and care taken in locating weaknesses in our arguments, and we are most thankful. Any remaining errors are of course our own responsibility.

## A. Intractability of Reasoning with ONE-OF

We present here a formal proof that subsumption with set descriptions is in fact NP-hard.[10]

We will show that if a term language allows a set description then it will need to do "case analysis" in order to check whether the extension of an individual belongs to a description or not; this is because this constructor behaves like disjunction if its elements can

10. Our original result was submitted for publication in 1990. A different, independent, proof of the same result has since been outlined in (Lenzerini & Schaerf, 1991).

be extensions of individuals whose membership in all terms is not known a priori, i.e., non-host individuals. In particular, we will show how to encode the testing of *unsatisfiability* of a formula in 3CNF as the question of recognizing an individual as an instance of a description. Since this problem is known to be NP-hard, we have strong indication of its intractability.

Start with a formula $F$, in 3CNF. Using DeMorgan's laws, construct formula $G$, which is the negation of $F$, and which is in 3DNF. Testing the validity of $G$ is equivalent to checking the unsatisfiability of $F$.

Construct for every propositional symbol $p$ used in $F$, two individual names P and $\hat{\mathsf{P}}$. (Here $\hat{\mathsf{P}}$ will represent the negation of $p$.) Each individual will have attribute truthValue, with possible fillers True and False

$$\mathsf{P}, \hat{\mathsf{P}} \quad \in \quad \forall \mathsf{truthValue}{:}\{\mathsf{True\ False}\}.$$

To make sure that P and $\hat{\mathsf{P}}$ have exactly one, and opposite, truth values, we create two more individual names, Yesp and Nop, with additional attributes approve and deny respectively, whose fillers need to have truth value True and False respectively:

$$\begin{array}{lcl} \mathsf{Yesp} & \in & \forall \mathsf{approve}{:}(\{\mathsf{P}\ \hat{\mathsf{P}}\} \sqcap \forall \mathsf{truthValue}{:}\{\mathsf{True}\}) \\ \mathsf{Nop} & \in & \forall \mathsf{deny}{:}(\{\mathsf{P}\ \hat{\mathsf{P}}\} \sqcap \forall \mathsf{truthValue}{:}\{\mathsf{False}\}) \end{array}$$

Now, given the formula $G = C1 \vee C2 \vee \ldots \vee Cn$, create individual names C1, C2, ..., Cn, each with role conjuncts containing the propositions that are its conjuncts. For example, if $C1 = p \wedge \neg q \wedge \neg r$ then

$$\mathsf{C1} \quad \in \quad \forall \mathsf{conjuncts}{:}\{\mathsf{P}\ \hat{\mathsf{Q}}\ \hat{\mathsf{R}}\} \sqcap \geq 3\, \mathsf{conjuncts}.$$

Finally, construct individual G to have C1, C2, ..., Cn as possible fillers for a new role disjunctsHolding:

$$\mathsf{G} \quad \in \quad \forall \mathsf{disjunctsHolding}{:}\{\mathsf{C1\ C2} \ldots \mathsf{Cn}\}.$$

The formula $G$ will then be valid iff there is always at least one disjunct that holds. This is equivalent to membership in the concept VALID-FORMULAE defined as

$$\geq 1\, \mathsf{disjunctsHolding} \sqcap \forall \mathsf{disjunctsHolding}{:}(\forall \mathsf{conjuncts}{:}(\forall \mathsf{truthValue}{:}\{\mathsf{True}\})).$$

The above shows that recognizing whether individuals are instances of descriptions is intractable in the presence of set descriptions, minimum number restrictions, and value restrictions.

We can convert this into a question concerning the subsumption of two descriptions by essentially making all the individuals involved attribute-fillers for new dummy attributes, and their descriptions as restrictions on these attributes. Then if the description is non-empty then these attribute values must satisfy the corresponding restrictions.

So, define concept UPPER to be

$$\forall \mathsf{formula}{:}\mathsf{VALID\text{-}FORMULAE}$$

and define concept LOWER to be

$\forall$dummy1-p:($\{$P$\}$ $\sqcap$ [$P'$*s concept descriptor*]) $\sqcap$
$\forall$dummy2-p:($\{\hat{\text{P}}\}$ $\sqcap$ [$\hat{P}'$*s concept descriptor*]) $\sqcap$
$\forall$dummy3-p:($\{$Yesp$\}$ $\sqcap \ldots$) $\sqcap$
$\forall$dummy4-p:($\{$Nop$\}$ $\sqcap \ldots$) $\sqcap$
...
$\forall$dummy5-$c_i$:($\{C_i\}$ $\sqcap \ldots$) $\sqcap$
...
$\forall$formula:($\{$G$\}$ $\sqcap \ldots$)

Then in any database state either concept LOWER has no instances, in which case it is a subset of the extension of UPPER, or it has at least one instance, in which case the individual names filling the various dummy attributes must have the properties ascribed to them, whence C will be in VALID-FORMULAE (and hence UPPER will subsume LOWER) iff C is valid, which completes the proof.